%% file: main.tex
\documentclass[10pt,twocolumn,letterpaper]{article}

\usepackage[pagenumbers]{cvpr} 


\definecolor{cvprblue}{rgb}{0.21,0.49,0.74}
\usepackage[pagebackref,breaklinks,colorlinks,allcolors=cvprblue]{hyperref}
\usepackage{tcolorbox}
\usepackage{makecell}
\usepackage{tabularx}
\usepackage[accsupp]{axessibility}  

\tcbset{
    mypromptbox/.style={
        colback=gray!5!white,      
        colframe=lightgray,             
        width=\linewidth,          
        boxrule=0.5mm,             
        arc=1mm,                   
        outer arc=1mm,             
        fonttitle=\bfseries,       
        coltitle=black,            
    }
}
\usepackage{listings}

\usepackage{lipsum}
\newcommand\blfootnote[1]{%
  \begingroup
  \renewcommand\thefootnote{}\footnote{#1}%
  \addtocounter{footnote}{-1}%
  \endgroup
}

\def\paperID{16778} 
\def\confName{CVPR}
\def\confYear{2025}

\title{FactCheXcker: Mitigating Measurement Hallucinations \\ in Chest X-ray Report Generation Models}

\author{
    Alice Heiman$^{1*}$ \quad
    Xiaoman Zhang, PhD$^{2*}$ \quad
    Emma Chen, MS$^2$ \\
    Sung Eun Kim, MD$^2$ \quad
    Pranav Rajpurkar, PhD$^{2\dagger}$  \\
$^1$Stanford University, USA
\quad $^2$ Harvard University, USA\\
}

\begin{document}
\maketitle

\blfootnote{$^*$ Equal contribution. \quad $^\dagger$ Corresponding author.}

\input{sec/0_abstract}    
\input{sec/1_introduction}

\input{sec/2_related_work}

\input{sec/3_method}

\input{sec/4_experiments}

\input{sec/5_results}

\input{sec/6_discussion}

{
    \small
    \bibliographystyle{ieeenat_fullname}
    \bibliography{main}
}

\input{sec/X_suppl}

\end{document}

%% file: sec/0_abstract.tex
\begin{abstract}
Medical vision-language models often struggle with generating accurate quantitative measurements in radiology reports, leading to hallucinations that undermine clinical reliability. We introduce FactCheXcker, a modular framework that de-hallucinates radiology report measurements by leveraging an improved query-code-update paradigm. Specifically, FactCheXcker employs specialized modules and the code generation capabilities of large language models to solve measurement queries generated based on the original report.
After extracting measurable findings, the results are incorporated into an updated report. We evaluate FactCheXcker on endotracheal tube placement, which accounts for an average of 78\% of report measurements, using the MIMIC-CXR dataset and 11 medical report-generation models. 
Our results show that FactCheXcker significantly reduces hallucinations, improves measurement precision, and maintains the quality of the original reports. 
Specifically, FactCheXcker improves the performance of 10/11 models and 
achieves an average improvement of 135.0\% in reducing measurement hallucinations measured by mean absolute error. Code is available at \url{https://github.com/rajpurkarlab/FactCheXcker}.

\end{abstract}





%% file: sec/1_introduction.tex
\section{Introduction}
\label{sec:intro}

\begin{figure}[!t]
    \centering
    \includegraphics[width=1\linewidth]{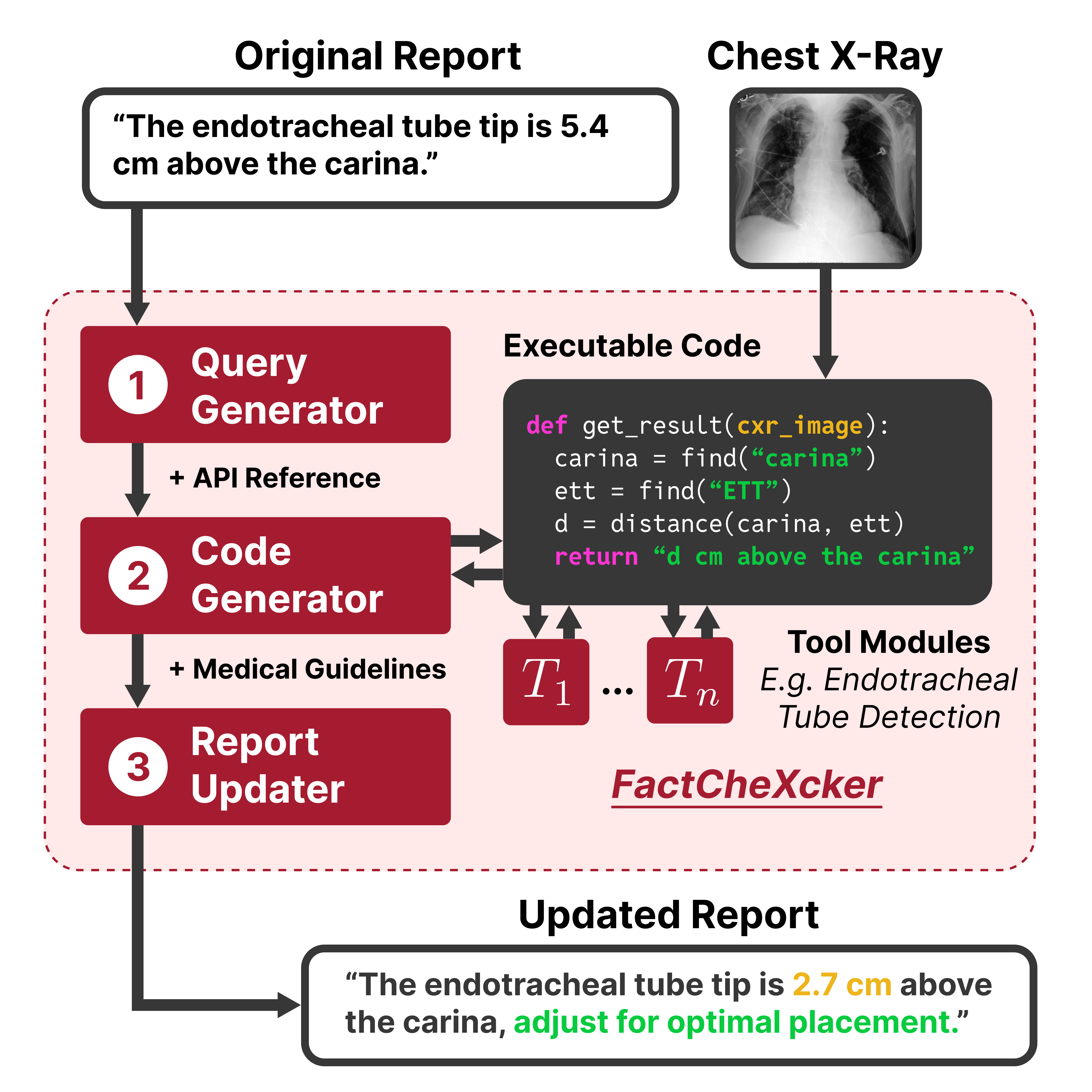}
    \caption{An illustration of \textbf{FactCheXcker} pipeline, which uses a query-code-update approach to alleviate measurement hallucinations in chest X-ray report generation models.}
    \label{fig:pipeline}
\end{figure}

Medical vision-language models have shown significant potential in assisting with everyday clinical tasks, such as interpreting chest X-rays for radiology report generation~\cite{sloan2024automated,zhou2024generalist,bannur2024maira,hyland2023maira}.
Recent surveys of radiologists' expectations regarding AI systems in clinical practice have revealed a preference for medical models on more functional, tedious tasks such as calculating medical scores and assessing organ sizes, volumes, or density over general-purpose interpreters~\cite{yildirim2024multimodal}.
However, current medical report generation models often struggle with such quantitative measurements, including determining the size of a lung nodule or measuring the distance from an endotracheal tube (ETT) to the carina~\cite{zhang2024uncovering}.
Incorrect or missing measurements can lead to adverse clinical outcomes since many reporting guidelines rely on precise thresholds.
This phenomenon, usually referred to as ``hallucination'' in the context of AI models, limits their reliability and potential for clinical deployment.

These hallucinations primarily stem from the inherent complexity of tasks that require precise measurements.
Effective measurement requires a model to excel at multiple subtasks simultaneously: identifying measurable findings, confirming their presence in images, localizing these findings, pinpointing crucial anatomical landmarks, and performing accurate dimensional calculations.
Without explicit instruction or specialized architectures for quantitative reasoning, most current report generation models struggle to achieve the precision required for clinical measurements. 
However, enhancing these capabilities becomes more feasible by integrating existing specialized vision models for each specific task, thus leveraging their expertise to improve overall measurement accuracy.


\begin{figure}[!t]
    \centering
    \includegraphics[width=1\linewidth]{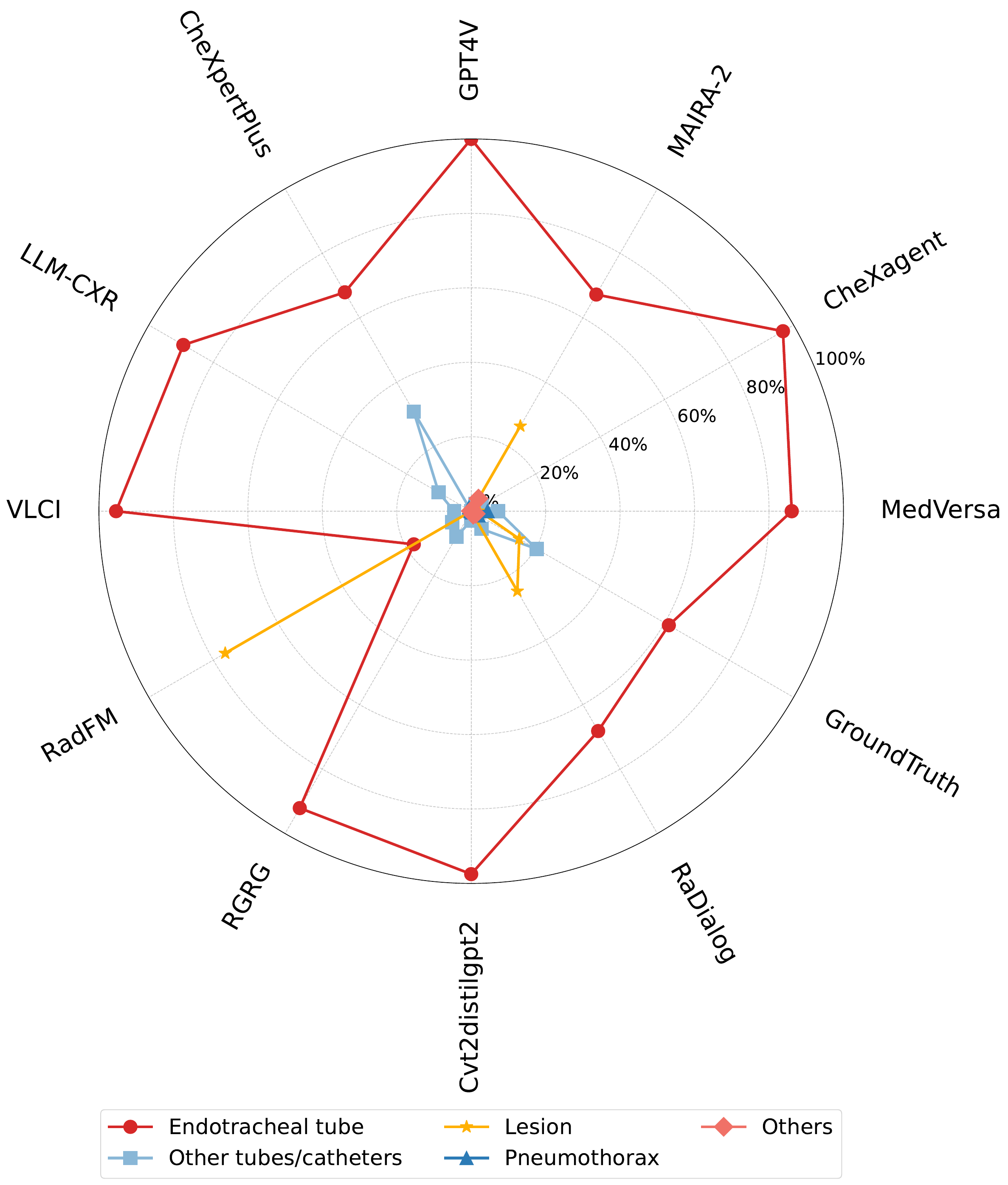}
    \caption{Distribution of measurable findings across different models. The radial axis shows the percentage of each finding type relative to the total measurable findings for each model. Notably, endotracheal tubes (shown in red) represent, on average, 78\% of measurements across all models. The remaining categories account for an average of 12.1\%, 8.4\%, 0.7\%, and 0.5\% of measurements, respectively.}
    \label{fig:counts}
\end{figure}

Motivated by this, we present \textbf{FactCheXcker}. This extensible tool-based pipeline reevaluates and updates measurements in model-generated radiology reports without retraining or modifying the original model. FactCheXcker follows a query-code-update paradigm.
Specifically, the framework automatically generates measurement queries based on the original radiology report, such as ``measure [XYZ]''. It generates executable code based on a domain-specific API that solves the measurement query and returns a result. 
Furthermore, we extend the pipeline to automatically incorporate the results into an updated report. 
Thus, the FactCheXcker pipeline can iteratively improve any radiology report without retraining the original model. 
The flexible format allows FactCheXcker to solve a wide range of measurement queries by combining modular methods to perform more complex tasks, such as finding the distance between two anatomies or counting the number of nodules.



\begin{table}[]
\centering
\small
\caption{Examples of specific measurements of various objects in radiology reports.}
\vspace{-5pt}
\renewcommand{\arraystretch}{1}
\begin{tabular}{l p{5cm}}
\toprule
Measurement & Example \\
\midrule
Endotracheal Tube & Endotracheal tube tip measures approximately 4.3 cm above the carina. \\
Lesion & The lesion is larger since the prior examination where it measured 11 mm. \\
Other tubes/catheters & A right PICC has its tip terminating in the proximal right atrium, which should be retracted 2 cm. \\
Pneumothorax & Moderate right apical pneumothorax measuring 2.3 cm at the apex. \\
Others & The balloon pump lies 2.3 cm from the apex of the aortic arch. \\
\bottomrule
\end{tabular}
\label{tab:medical_measurements}
\vspace{-8pt}
\end{table}

Our investigation begins with analyzing the most common measurable findings mentioned in radiology reports across 11 report generation models and real-world data. Figure \ref{fig:counts} shows that endotracheal tubes comprise most concrete measurements. ETTs assist patients with mechanical ventilation but require regular monitoring through chest X-rays to ensure proper positioning, thereby minimizing the risk of faulty intubations. Accurate placement is critical, as an improperly positioned ETT can result in complications such as hypoxia, pneumothorax, or even death~\cite{hagberg2005complications}. 
Radiologists typically measure the distance between the carina and the tip of the ETT in centimeters and recommend repositioning if it is outside the optimal range.
Precise measurement is, therefore, crucial to ensuring proper patient care. 
Table~\ref{tab:medical_measurements} lists measurements seen in radiology reports.

To evaluate the effectiveness of FactCheXcker, we apply it to the task of assessing ETT placement. We evaluate FactCheXcker on the MIMIC-CXR dataset~\cite{johnson2019mimic} for the ETT placement task using 11 different report-generation models. Our results demonstrate that FactCheXcker significantly reduces measurement hallucinations and improves ETT detection and placement precision, all while preserving the overall quality of the original reports. In summary, our main contributions are as follows:
\begin{itemize}
    \item We propose FactCheXcker, an extensible and modular framework to mitigate measurement hallucinations in chest X-ray reports.
    \item We evaluate FactCheXcker on the task of endotracheal tube placement, a crucial part of chest X-ray interpretation in intensive care units, demonstrating its ability to reduce false positives and improve measurement accuracy.
    \item We show that applying FactCheXcker improves the performance of 10/11 models and significantly reduces hallucinated ETT placement measurements, achieving an average improvement of 135\% in mean absolute error.
\end{itemize}

%% file: sec/2_related_work.tex
\section{Related Work}
\label{sec:related}

\vspace{3pt} \noindent \textbf{Medical Report Generation Models.}
Medical report generation has seen significant advances with the emergence of large language models. 
Early approaches utilized CNN-RNN architectures to generate reports from medical images~\cite{sirshar2022attention,xue2018multimodal}. At the same time, more recent works have adopted transformer-based architectures for improved performance~\cite{chen2023cross,tanida2023interactive,lee2023llm,tu2024towards}. 
These models typically combine vision encoders, such as Vision Transformer~\cite{dosovitskiy2020image} and Swin Transformer~\cite{liu2021swin}, with text decoders to generate detailed clinical descriptions. 
For instance, CheXpertPlus baseline models~\cite{chambon2024chexpert} integrates Swinv2~\cite{liu2022swin} architecture with a two-layer BERT decoder~\cite{kenton2019bert} for medical report generation. 
In addition, more versatile foundation models have emerged, such as MedVersa~\cite{zhou2024generalist} which coordinates multimodal inputs and outputs for varying tasks, RadFM~\cite{wu2023towards} which supports both 2D and 3D scans with visual-language interleaving, and CheXagent~\cite{chen2024chexagent} which takes an instruction-tuned approach with a specialized vision encoder bridged with language modalities. 
Recently, MAIRA-2~\cite{bannur2024maira} combines radiology-specific image encoders with large language models (LLMs) for grounded report generation. 
Despite these architectural advancements, critical limitations remain: the models often struggle with accurately interpreting fine-grained quantitative information and spatial relationships within medical images, highlighting the need for specialized approaches.


\vspace{3pt} \noindent \textbf{Enhancing LLM Capabilities with External APIs.}
Integrating LLMs with external APIs has become popular for expanding action spaces and tackling complex tasks~\cite{schick2024toolformer,liang2024taskmatrix,li2023api}. For instance, HuggingGPT~\cite{shen2024hugginggpt} utilizes models from the HuggingFace platform to dynamically call specialized models and address diverse user needs. Similarly, ViperGPT~\cite{suris2023vipergpt} uses Codex, an LLM designed to convert text descriptions into executable Python code, allowing the model to complete tasks by directly running generated code.

\vspace{3pt} \noindent \textbf{Fact-Checking and Mitigating Hallucination in LLMs.}
Hallucinations represent a significant challenge in LLMs, where the models generate content that is not grounded in the input signal but instead relies on their internal knowledge~\cite{ji2023survey}. 
This issue has increased focus on fact-checking and mitigating hallucinations across multiple modalities~\cite{gunjal2024detecting,liu2023mitigating,bai2024hallucination}. 
One line of research to address this problem has focused on hallucination detection frameworks and fact-checking mechanisms for LLM outputs. 
FacTool~\cite{chern2023factool} proposes a task- and domain-agnostic framework that leverages GPT for extracting and verifying claims. 
ProgramFC~\cite{pan2023fact} and FActScore~\cite{min2023factscore} both approached complex fact-checking by breaking down statements into simpler subtasks or atomic statements that could be individually verified against a knowledge source. 
In multimodal LLMs, approaches~\cite{li2023evaluating,lovenia2023negative,wang2024mitigating} have been proposed that include using binary questions to assess object presence in visual inputs or leveraging Visual Question Answering (VQA) methods to generate negative statements indicating the absence of particular objects in visual scenes.

Despite progress, a significant gap remains in the medical domain for real-life clinical usage. Medical imaging requires precise measurements, such as identifying the size of lesions or the exact positioning of medical devices. Inspired by these works, we propose leveraging LLMs and APIs to mitigate measurement hallucination. Our method uses specialized medical tools for accurate anatomical calculations, and automatically updates verified measurements in the report. This approach has two main advantages: no assumption of model weight access and portability to resource-limited and/or privacy-centered hospital settings. Small and locally deployable tool modules can serve multiple models while maintaining data compliance, avoiding the need for extensive model tuning or external data transfer.

%% file: sec/3_method.tex
\section{Method}
\label{sec:methods}

In this section, we introduce \textbf{FactCheXcker}\footnote{\url{https://github.com/rajpurkarlab/FactCheXcker}}, an extensible and modularized pipeline designed to eliminate hallucinated measurements in radiology reports. As illustrated in Figure~\ref{fig:pipeline}, FactCheXcker consists of three main components: Query Generator, Code Generator, and Report Updater. 
When provided with a medical image and its corresponding model-generated report that may contain hallucinated measurements, the Query Generator identifies potential measurement discrepancies in the report, the Code Generator creates and executes specialized code to obtain precise measurements from the image, and the Report Updater integrates these verified measurements into the report.

\subsection{Task Definition}
We define hallucination as making an erroneous measurement of the presence of a tube when it does not exist or making the wrong measurement. We categorize hallucinations in medical report generation models into two types: object hallucination and measurement hallucination.
Object hallucination occurs when the model erroneously predicts the presence of specific diseases in particular anatomical structures, such as incorrectly identifying pneumonia in the right lung lobe. On the other hand, measurement hallucination refers to inaccurate predictions of numerical values related to quantifiable findings, such as the size of lesions or distances between devices and anatomical structures.
The extensible design of FactCheXcker allows it to address both categories of hallucinations. However, in this paper, we primarily focus on detecting and correcting measurement hallucinations in model-generated radiology reports, as this issue is more prevalent and critically requires our approach.

\subsection{FactCheXcker Framework}

\vspace{0.3cm}\noindent\textbf{Query Generator.} 
This module receives the original report and infers what information needs to be re-measured and verified. 
It then generates a list of measurement queries that guide subsequent steps. For instance, if the original report was ``The endotracheal tube terminates 2.3 cm above the carina. There is a new dense right central opacity measuring about 6 cm x 3 cm''; the framework would produce the queries ``measure the distance between the endotracheal tube and the carina" and ``measure the dimensions of the right central opacity''. For this purpose, we utilize GPT-4o mini~\cite{hurst2024gpt} as the query generator, and details on the prompt used can be found in the Supplementary Material. These queries are then passed on to the Code Generator module.

\vspace{0.3cm}\noindent\textbf{Code Generator.}
This module receives the measurement query and a high-level API description. It dynamically generates Python code to answer the query based on the methods provided by the API. 
We have designed a comprehensive toolbox for identifying and correcting measurement hallucinations in radiology reports, as detailed below:
 

\begin{itemize}
    \item \textbf{exists(\textbf{object})}: Returns True if the object is found in the image, and False otherwise.
    \item \textbf{find(object)}: Returns a list of the bounding box or segmentation map of the matched objects in the image.
    \item \textbf{distance(object\_a, object\_b)}: Returns the distance (in cm) between the center of two objects in the image.
    \item \textbf{width(object)}:  Returns the largest width (in cm) of the object.
    \item \textbf{height(object)}:  Returns the largest height (in cm) of the object.
    \item \textbf{diameter(object)}:  Returns the largest diameter (in cm) of the object, namely \textsc{max(width, height)}.
    \item \textbf{dimensions(object)}: Returns the dimensions (in cm) of the object according to major axis x minor axis.
    \item \textbf{within(object\_a, object\_b)}: Returns True if the object\_a (typically a specific finding)'s center is within the object\_b (typically an anatomical structure)'s bounding box or segmentation map, False otherwise.
\end{itemize}
Each method is implemented using specialized tool modules that can be swapped out without changing the API interface. We use a mixture of plug-and-play open-source and fine-tuned models. See \textit{Experiments} for the specific tools used for the results in this paper.

\vspace{0.3cm}\noindent\textbf{Report Updater.}
\noindent This module takes the original report together with all the results of the measurement queries to update the report. Specifically, it updates a sentence if a new value is presented, or removes a phrase entirely if the object was hallucinated. 
Moreover, since the old interpretation may be outdated in light of new evidence, the module references measurement guidelines for each result to determine if the ETT placement should be updated. The pipeline concludes with returning the updated report.
 For instance, in the case of ETT placement, where the ideal position is 5 ± 2 cm above the carina ~\cite{goodmanRadiographicEvaluationEndotracheal1976}, if the updated measurement falls outside this range, the Report Updater will flag this discrepancy and update the report to indicate the need for repositioning. We notice that the numeric reasoning of GPT-4o mini~\cite{hurst2024gpt} is currently insufficient to reliably extract the inference. For instance, the language model fails to identify if a measurement is below or above a threshold value. Therefore, in this version of the pipeline, we implement a strict rule-based approach to infer the placement category of the endotracheal tube. If the distance is between 3-7 centimeters, the placement is deemed as ``correct''. Otherwise, the placement is updated to ``incorrect''.

%% file: sec/4_experiments.tex
\section{Experiments}
\label{sec:experiments}

\subsection{Datasets and Tasks}

\noindent\textbf{Datasets.} We utilize MIMIC-CXR 2.0.0~\cite{johnson2019mimic}, a public and de-identified chest X-ray dataset containing 377,110 images from 227,835 radiographic studies performed at the Beth Israel Deaconess Medical Center. Each study includes both the radiographic images and corresponding free-text radiology reports. 
For pipeline development, we use the training- and validation sets. For the final evaluation and analysis, we use the test set, which comprises 2288 studies, including 182 cases with ETT mentions.

\vspace{3pt}\noindent\textbf{Tasks.} 
We evaluate our approach on a critical clinical task: Endotracheal Tube Placement. Accurate measurement of these tasks is essential for ensuring patient safety and facilitating timely clinical interventions. 
To delve deeper into the endotracheal tube positions, we categorize the mention of endotracheal tubes into three categories: \textit{presence}, \textit{measurement}, and \textit{placement}. Presence is a binary classification that is \textit{True} if the report mentions that the image contains an endotracheal tube, and \textit{False} otherwise. Measurement is a real-valued number that is the distance between the endotracheal tube tip and the carina. The placement category is the inference made about the position – namely if the endotracheal tube position is correct, or incorrect (either too high or too low) and thus requires attendance.

\subsection{Tool Modules}
\noindent Here are the experiment details of the tools used for the API methods in this paper.
\begin{itemize}
    \item \noindent\textbf{exists(``endotracheal tube'')}: We fine-tune a ResNet-50 model~\cite{heDeepResidualLearning2015} pre-trained on ImageNet~\cite{deng2009imagenet} on the task of endotracheal tube binary classification, which we call \textsc{ResNet-50+}. We extract 5,000 negative and 5,000 positive samples from the MIMIC-CXR train dataset using GPT-4o mini based on mentions of endotracheal tubes in the ground truth report. We train using PyTorch Lightning, for 30 epochs, with OneCycleLR~\cite{smithSuperConvergenceVeryFast2018}. See Table \ref{tab:resnet_performance} for the performance on the MIMIC-CXR test set.
    \item \noindent\textbf{find(``carina'', ``endotracheal tube'')}: We fine-tune CarinaNet~\cite{oliverImageAugmentationAutomated2023} on a diverse private dataset from 22 hospitals. The CarinaNet module, which we call \textsc{CarinaNet+}, outputs an (x, y)-coordinate for both the carina and the endotracheal tube along with confidence scores. See Table \ref{tab:carinanet_performance} for the MIMIC-CXR test set performance.
    \item \noindent\textbf{find(``heart'', ``left lung'', etc.)}: We use the ChestX-Det Segmentation Model from TorchXRayVision~\cite{cohenTorchXRayVisionLibraryChest2021} to generate segmentation maps for 14 anatomical regions, such as the heart, lungs, mediastinum, and diaphragm. 
    \item \noindent\textbf{distance(), diameter(), dimensions()}: We use the pixel spacing metadata of the MIMIC-CXR images as well as the original- and model-specific image dimensions to convert pixel distances into physical distances in cm.
\end{itemize}

\begin{table}[]
\centering
\small
\caption{Performance of the \textsc{CarinaNet+} Module for the function \textbf{find(``endotracheal tube'', ``carina'')}.}
\vspace{-5pt}
\begin{tabular}{lcccccc}
\toprule
Module & MAE & MSE & Max & Min & Avg & Std \\
\midrule
CarinaNet+ & 0.94 & 2.57 & 12.15 & 0.00 & 0.94 & 1.30 \\\bottomrule
\end{tabular} 
\label{tab:carinanet_performance}
\end{table}

\begin{table}[]
\centering
\small
\caption{Performance of the \textsc{ResNet-50+} Module for the function \textbf{exists(``endotracheal tube'')}.}
\vspace{-5pt}
\begin{tabular}{lcccccc}
\toprule
Module & ACC & BACC & F1 & Prec. & Rec. & AUC \\
\midrule
ResNet-50+ & 0.94 & 0.97 & 0.73 & 0.57 & 0.99 & 0.97 \\\bottomrule
\end{tabular}
\label{tab:resnet_performance}
\end{table}

\subsection{Baselines and Implementation Details}

\vspace{3pt}\noindent\textbf{Baseline Models.}
We demonstrate FactCheXcker's effectiveness across 11 different medical report generation models, including CheXagent~\cite{chen2024chexagent}, CheXpertPlus~\cite{chambon2024chexpert}, 
GPT-4V~\cite{yang2023dawn}, 
LLM-CXR~\cite{lee2023llm}, 
MAIRA-2~\cite{bannur2024maira}, MedVersa~\cite{zhou2024generalist}, RadFM~\cite{wu2023towards}, RaDialog~\cite{pellegrini2023radialog}, RGRG~\cite{tanida2023interactive}, 
VLCI~\cite{chen2023cross}.  
For all models, we follow the official protocols to generate reports on the MIMIC-CXR test set.
Detailed descriptions and implementation specifications are provided in the Supplementary Material.

\vspace{3pt}\noindent\textbf{Measurement Extraction.}
We implement a systematic extraction process to extract labels of the concrete measurements made in the ground truth- and model-generated reports.
To optimize API usage, we first filter reports containing specific measurements using keywords: ``cm'', ``mm'', ``centimeter(s)'', ``millimeter(s)'', ``measure(s)''. 
We then utilize GPT-4o mini to extract object names with associated concrete measurements from both ground truth and model-generated reports. The detailed extraction prompt is provided in the Supplementary Material. We categorize the identified measurements into five main categories: Endotracheal tube, Other tubes/catheters (e.g., PICC, central venous line), Lesion (e.g., opacity, mass, nodule), Pneumothorax, and Others (e.g., calcification, balloon pump). Notably, endotracheal tubes account for the most significant proportion of measurements, with an average mention percentage of 78.3\% across all models. 

\vspace{3pt}
\noindent\textbf{Endotracheal Tube Annotation.} 
After extraction of the measurements from the MIMIC-test set, there remained 45 reports where the ground truth report mentioned an endotracheal tube as present, but did not give a concrete measurement. To give more opportunity for comparisons, a trained radiologist annotated the location of the carina and the endotracheal tube tip in the 45 images.
Then, we computed the distance between the centers of the annotated bounding boxes as the distance between the carina and the endotracheal tube. 
This result was passed to the Report Updater-module of our framework, and thus the result was ``injected'' into the ground truth reports for later comparisons.


\begin{table*}[!t]
\centering
\small
\setlength{\tabcolsep}{5pt}
\caption{Performance Comparison between the Original Model and the Updated Model with FactCheXcker in the tasks of ETT Presence, Measurement and Placement. For Precision metrics, higher values indicate better performance. For MAE and Composite metrics, lower values indicate better performance. The Improvement percentage shows the relative performance gain. \textbf{Bold} values indicate better performance between Original and Updated models for each metric.}
\begin{tabular}{lcccccccccc}
\toprule
& \multicolumn{2}{c}{ETT Presense} & \multicolumn{6}{c}{ETT Measurement} & \multicolumn{2}{c}{ETT Placement} \\
\cmidrule(lr){2-3} \cmidrule(lr){4-9} \cmidrule(lr){10-11}
Model & \multicolumn{2}{c}{Precision} & \multicolumn{3}{c}{MAE} & \multicolumn{3}{c}{Composite} & \multicolumn{2}{c}{Precision} \\
\cmidrule(lr){2-3} \cmidrule(lr){4-6} \cmidrule(lr){7-9} \cmidrule(lr){10-11}
 & Original & Updated & Original & Updated & Improvement & Original & Updated & Improvement & Original & Updated \\
\midrule
CheXagent & 0.70 & \textbf{0.76} & 1.98 & \textbf{0.68} & 191.0\% & 4.12 & \textbf{1.39} & 196.0\% & 0.85 & \textbf{0.90} \\
CheXpertPlus & 0.64 & \textbf{0.68} & 1.44 & \textbf{0.66} & 118.0\% & 2.09 & \textbf{0.92} & 127.0\% & 0.90 & \textbf{0.94} \\
Cvt2distilgpt2 & 0.71 & \textbf{0.72} & 3.82 & \textbf{0.89} & 329.0\% & 6.37 & \textbf{1.46} & 336.0\% & 0.73 & \textbf{0.88} \\
GPT4V & 0.20 & \textbf{0.58} & 2.35 & \textbf{0.39} & 503.0\% & 11.19 & \textbf{1.22} & 817.0\% & \textbf{1.00} & \textbf{1.00} \\
LLM-CXR & 0.08 & \textbf{0.58} & 2.41 & \textbf{0.91} & 165.0\% & 18.54 & \textbf{1.65} & 1024.0\% & 0.74 & \textbf{0.97} \\
MAIRA-2 & 0.63 & \textbf{0.68} & 1.43 & \textbf{0.99} & 44.0\% & 2.01 & \textbf{1.34} & 50.0\% & 0.76 & \textbf{0.91} \\
MedVersa & 0.73 & \textbf{0.80} & 1.07 & \textbf{0.86} & 24.0\% & 1.49 & \textbf{1.13} & 32.0\% & 0.84 & \textbf{0.94} \\
RGRG & 0.50 & \textbf{0.60} & 1.58 & \textbf{0.76} & 108.0\% & 2.51 & \textbf{1.09} & 130.0\% & 0.77 & \textbf{0.96} \\
RaDialog & 0.67 & \textbf{0.69} & 0.99 & \textbf{0.86} & 15.0\% & 1.43 & \textbf{1.23} & 16.0\% & 0.81 & \textbf{0.92} \\
RadFM & 0.28 & \textbf{0.47} & \textbf{0.65} & 1.04 & -38.0\% & \textbf{8.12} & 11.56 & -30.0\% & \textbf{1.00} & \textbf{1.00} \\
VLCI & 0.25 & \textbf{0.54} & 3.50 & \textbf{0.98} & 257.0\% & 21.88 & \textbf{4.90} & 347.0\% & 0.80 & \textbf{0.88} \\
\midrule
Average & 0.49 & \textbf{0.65} & 1.93 & \textbf{0.82} & 135.0\% & 7.25 & \textbf{2.54} & 186.0\% & 0.84 & \textbf{0.94} \\
\bottomrule
\end{tabular}
\label{tab:ett_presence_comparison}
\end{table*}

\subsection{Evaluation Metrics}

Previous work has introduced several metrics for measuring hallucinations in image captioning, such as CHAIR~\cite{rohrbach2018object} and POPE~\cite{li2023evaluating}. However, CHAIR assumes complete ground truth labels and only handles binary hallucinations and POPE requires direct model querying which is unavailable in our setup. Furthermore, existing metrics place less emphasis on false negatives, which are critical in medical contexts where missing observations could be fatal. Thus, to comprehensively assess the performance of our model, we propose to evaluate it from three perspectives: presence (object) hallucination, measurement hallucination, and placement hallucination.


\vspace{3pt} \noindent \textbf{Presence Hallucination.} To evaluate hallucinations related to the identification of findings or devices, we employ Precision, Recall, F1-Score and Balanced Accuracy. These metrics measure the model's ability to identify the ETT presence accurately without introducing false mentions. Since the FactCheXcker pipeline does not address false positive rate, we focus on the precision score. All the result metrics are included in the Supplementary Material.

\vspace{3pt} \noindent \textbf{Measurement Hallucination.} To evaluate measurement hallucinations, we use the mean absolute error (MAE). Specifically, we compute the mean absolute error between samples with a ground truth measurement. A perfect error of zero means that the model correctly identified the presence of an ETT tube \textit{and} made the perfect measurement. If the ground truth contains a measurement but not the model, the model value is set to 0 to penalize missed ETT detections and measurements. Finally, we also introduce a ${Composite}$ metric that combines MAE and the F1-Score into one number. Specifically, we compute the ${Composite}$ score using the following formula:

\begin{align*}
    \textsc{Composite} = \frac{\textsc{MAE}}{\textsc{F1-Score}}.
\end{align*}

\noindent A \textbf{lower} ${Composite}$ score is \textbf{better}. For example, if the model makes all measurements perfectly, the ${Composite}$ score would be zero. If the F1-Score is perfect (namely an F1-Score of 1), the ${Composite}$ score becomes the MAE. However, any imperfect F1-Score would essentially become a multiplier of the MAE score.

\vspace{3pt} \noindent \textbf{Placement Hallucination.} In the current setting, we use a rule-based approach to interpret ETT placement as ``correct'' or ``incorrect'' based on ETT measurement. In cases where a model-generated report identifies an ETT but lacks details of its placement or measurement, we assume by default that the placement is deemed correct. We base this on the assumption that the report should explicitly mention any notable finding that needs attending. Then, we compute Precision, Recall, F1-Score, and Balanced Accuracy. The Supplementary Material includes all result metrics.







%% file: sec/5_results.tex
\section{Results}
\label{sec:results}


\subsection{Analysis of Measurement Hallucinations}
We first examine the distribution of measurable findings presented in Figure~\ref{fig:counts}, ETT measurements constitute a substantial portion of these findings, accounting for an average of 78\% of total measurements across model predictions and 61\% of measurements in the ground truth data. 
This high prevalence underscores the critical importance of accurate ETT measurements in clinical settings, as even minor positional errors can lead to severe complications in intensive care situations.
Thus, the occurrence of hallucinated ETT measurements in model-generated reports highlights a concerning trend, as these hallucinations could mislead critical interventions. Addressing these hallucinations is crucial for ensuring safe and effective management of intensive care interventions.

\begin{figure*}[htb]
    \centering
    \includegraphics[width=1\linewidth]{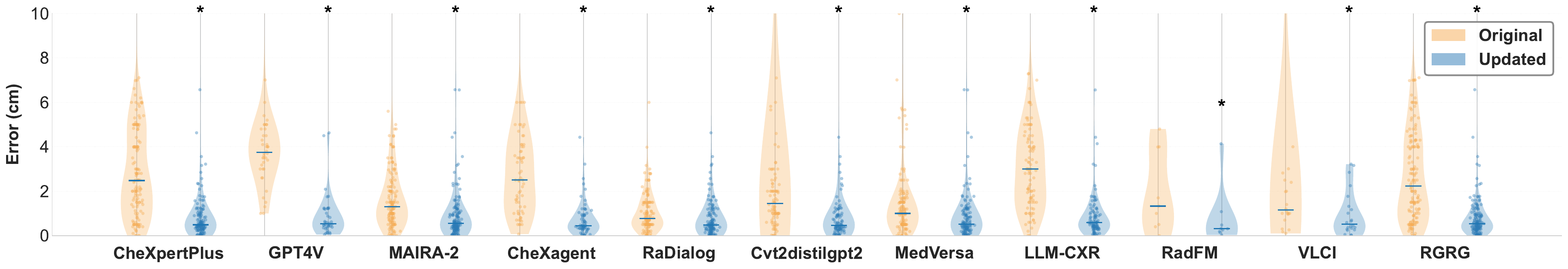} \vspace{-10pt}
    \caption{Distribution of absolute measurement errors for ETT placement across different models before and after using FactCheXcker. For each model, violin plots illustrate the error distribution, with scattered points representing individual cases. The blue median line in each violin plot indicates the central tendency, and errors are capped at 10 cm for visualization clarity. * indicates statistical significance with a p-value $<$ 0.1. Note that smaller measurement errors indicate better performance in ETT placement assessment.} \vspace{-10pt}
    \label{fig:accumulate_error}
\end{figure*}

\begin{figure*}[htb]
    \centering
    \includegraphics[width=1\linewidth]{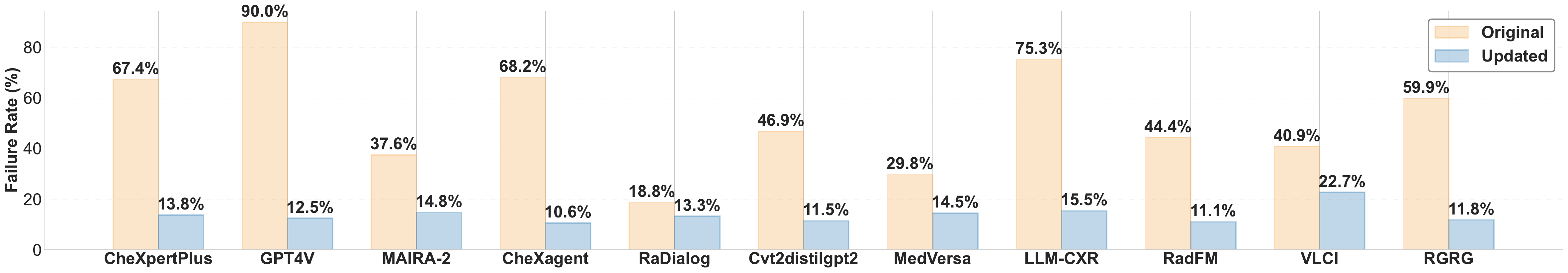} \vspace{-5pt}
    \caption{Comparison of failure rates for ETT placement across different models before and after using FactCheXcker. A case is considered failed if the ETT placement measurement error exceeds 1.5 cm. Note that smaller failure rates indicate better performance.}
    \label{fig:failed_case}
\end{figure*}

\begin{figure*}[!t]
    \centering
    \includegraphics[width=1\linewidth]{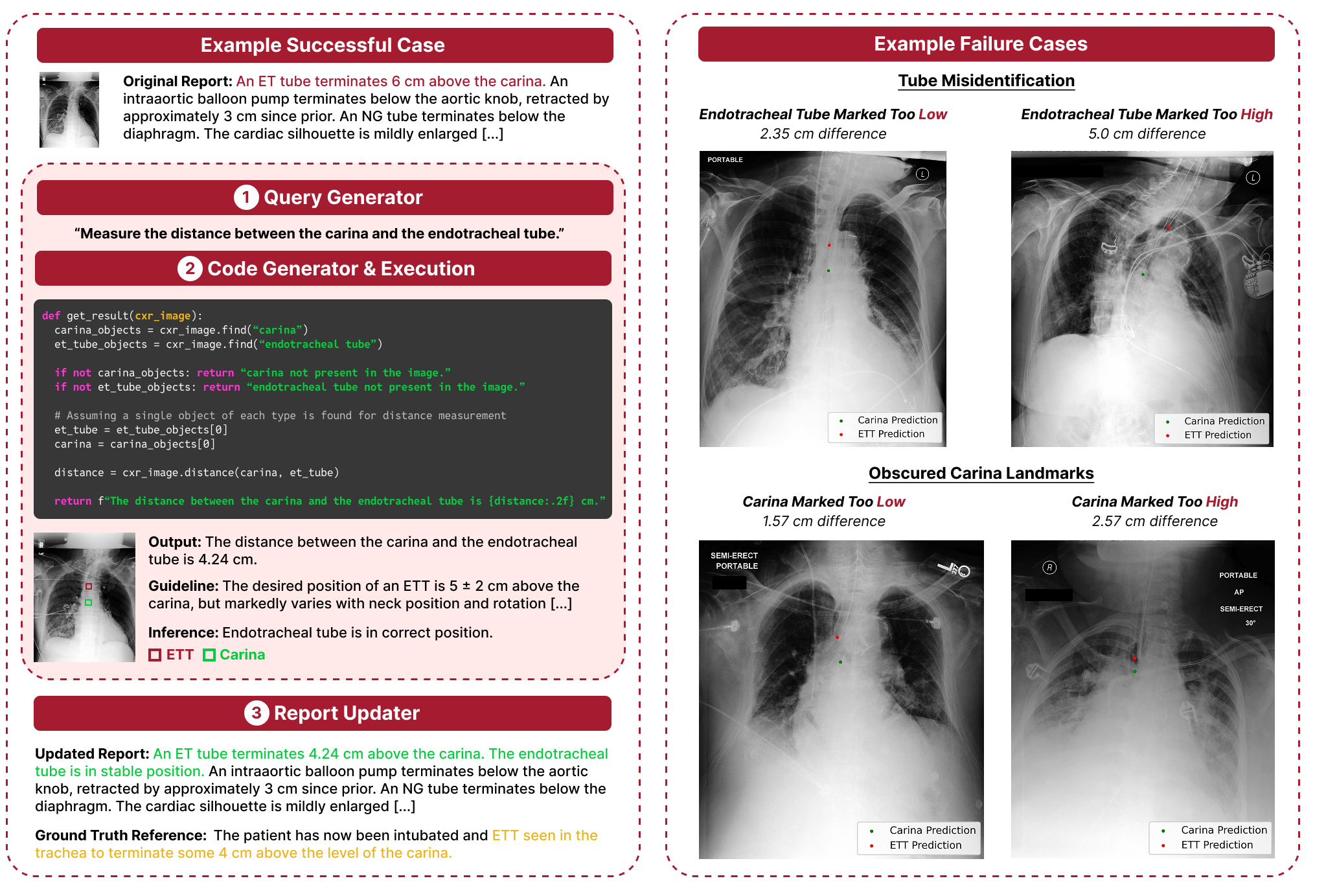}\vspace{-5pt} 
    \caption{\textbf{FactCheXcker Case Studies}. (Left) Given an input image and an initial model-generated report, FactCheXcker generates queries for the code generator, which synthesizes a program. This program is then executed using a Python interpreter to produce the output. The update module leverages this output to generate an updated report with accurate measurements. (Right) Failure cases include tube misidentifications and obscured carina landmarks.}  
    \label{fig:examples}
\end{figure*}

\subsection{FactCheXcker for ETT Presence}
We evaluate FactCheXcker's capability in improving ETT presence detection, which is a binary classification task determining whether an endotracheal tube is present in the chest X-ray image. Notably, we consider all false positives as object hallucinations, regardless if the model report contains an accompanied measurement. Table~\ref{tab:ett_presence_comparison} presents a comparison between the original and updated models on the Precision score. 
The results demonstrate that FactCheXcker consistently enhances the ETT detection performance across all models. 
Model LLM-CXR shows the most significant improvement in precision with an increase of 1024\%.
MedVersa and CheXagent achieved the highest precision scores after enhancement (0.80 and 0.76 respectively), demonstrating FactCheXcker's ability to further improve models with strong baseline performance.
These improvements indicate that FactCheXcker effectively enhances models' ability to accurately identify the presence of endotracheal tubes while reducing false detections.

Note that in our current pipeline configuration, edits are made to the original report only if it mentions an endotracheal tube. Consequently, our analysis focuses on precision rather than recall, as the pipeline is not designed to address false negatives. A potential approach to mitigate this limitation involves running FactCheXcker on every image, independent of the report mentions. The results in the Supplementary Materials show that although this approach improves false negatives, the precision score significantly drops. The false positive rate depends on the ``exists'' modules. Further refinement of these modules could enhance results by reducing false positives and improving accuracy.

\subsection{FactCheXcker for ETT Measurement} 
To evaluate FactCheXcker’s effectiveness in mitigating measurement hallucinations, we analyze the measurement error on the instances where ETT is mentioned in both the original model-generated report and the ground truth report.
As shown in Table~\ref{tab:ett_presence_comparison}, for all models, FactCheXcker achieves significant improvement in MAE score and the composite scoring that combines MAE and precision.  
FactCheXcker obtains an average improvement of 135\% in the MAE score and an average improvement of 186\% in the composite score.
Specifically, models like VLCI, Cvt2distilgpt2, and LLM-CXR show remarkable improvements.
The results demonstrate a significant reduction in measurement errors across all metrics, indicating FactCheXcker's effectiveness in improving measurement accuracy for both correctly and incorrectly identified ETT cases.

To better understand the improvements from FactCheXcker, Figure~\ref{fig:accumulate_error} visualizes the distribution of absolute measurement errors across all models. The violin plots reveal that FactCheXcker substantially reduces measurement errors, with most models showing significantly lower median values after enhancement. GPT4V and LLM-CXR demonstrate the most dramatic error reductions, with their distributions becoming more concentrated around more minor errors. However, it is possible in some cases for the framework to make the measurement worse due to the imperfect nature of the modules. Pipeline improvements to address this limitation include intelligent use of module predictions and confidence scores to mitigate or express uncertainty.

Figure~\ref{fig:failed_case} presents the failure rates across models, where a failure is defined as a measurement error exceeding 1.5 cm.
Across all models, FactCheXcker demonstrates a clear improvement, significantly reducing the number of failed cases—nearly halving the failures on average. 
FactCheXcker achieved the most dramatic reduction with GPT4V, lowering the failure rate from 77.5\% to 22.5\%. 
Notably, FactCheXcker’s impact was particularly strong for models with high initial failure rates (above 70\%), such as GPT4V, CheXagent, and LLM-CXR, reducing their failure rates to below 20\%.
This consistent reduction in error demonstrates FactCheXcker's effectiveness in mitigating measurement hallucinations across diverse model architectures, establishing it as a valuable tool for enhancing the reliability of quantitative assessments in medical imaging models.

\subsection{FactCheXcker for ETT Placement}
We evaluate FactCheXcker's effectiveness in assessing ETT placement correctness, which classifies whether an endotracheal tube is positioned correctly or requires adjustment due to being too high or too low. Table~\ref{tab:ett_presence_comparison} compares the performance between the original and updated models and shows significant improvements in placement precision for all models with FactCheXcker, with an average improvement in placement precision from 0.84 to 0.94. 
Notably, for the LLM-CXR model, FactCheXcker increases placement precision from 0.74 to 0.97. 

\subsection{Discussion}
\noindent \textbf{Case Studies. }In Figure~\ref{fig:examples} we present examples illustrating FactCheXcker's performance in a successful case and in four failure cases. Specifically, in the successful case, FactCheXcker detects and corrects the faulty ETT measurement from 6 cm to about 4 cm above the carina. Our radiologist team manually reviewed all failure cases (20 in total, 1.95\% of all measurements). They identified three failure modes: tube misidentification (60\%), obscured carina landmarks due to haziness (35\%), and rotated images (5\%). Training is needed for complex cases with multiple tubes.

\noindent \textbf{Future Development.} A promising enhancement would be the integration of a module to detect and segment lesions, opacities, and nodules. 
Thus, the framework would be able to answer queries such as ``measure the cardiac-thoracic ratio'', or ``get the diameter of the largest nodule in the posterior right lung base''.
Moving forward, we will continually integrate additional tool modules into the framework to support an expanding range of diagnostic tasks.

%% file: sec/6_discussion.tex
\section{Conclusion}
\label{sec:conclusion}
This study introduces the challenge of detecting and mitigating measurement hallucinations in generated radiology reports. We present FactCheXcker, an extensible and modularized pipeline for leveraging existing open-source or quickly trainable modules to solve specific quantified tasks without having to retrain the original report generation model. 
Our evaluations demonstrate that FactCheXcker significantly reduces the occurrence of hallucinations, enhances the precision of measurements, and preserves the integrity and readability of the original reports.

\newpage
\noindent \textbf{Acknowledgment.}
Part of this effort was supported by the Harvard Medical School–Seoul National University Hospital–Seoul National University College of Medicine Collaborative Research Program (HMS-SNUH-SNUCM Collaborative Research Award). This award is part of the Boston-Korea Innovative Research Project, funded by the Korea Health Industry Development Institute (KHIDI) and the Ministry of Health and Welfare, Republic of Korea (Grant No. RS-2024-00403047).

%% file: sec/X_suppl.tex
\clearpage

\onecolumn
\appendix

\begin{center}
    {\Large \textbf{FactCheXcker: Mitigating Measurement Hallucinations \\ in Chest X-ray Report Generation Models}} \\[0.5em]
    \vspace{0.5em}
    {\large Supplementary Material}
\end{center}

\renewcommand{\thesubsection}{\Alph{subsection}}  

\setcounter{page}{1}



\subsection{Prompts}

\begin{tcolorbox}[title=Generate Queries from Report Prompt, mypromptbox, verbatim]
\footnotesize
\begin{lstlisting}
You are an intelligent radiologist tasked with verifying key measurements in radiology reports. 

Task: Given a radiology report, your responsibility is to identify if it mentions any of 
[endotracheal tube] and formulate a measurement query to verify the results in the report.

**Examples**:

- "Measure the distance between the carina and the endotracheal tube."

- "Measure the diameter of the lung nodules in the upper left lung."

- "Are there any lung masses present in the image?"

**Output Requirements**:

- Make sure the object is present in the image (positive examples). "No pneumothorax" and
"The patient has been extubated" are two negative examples.

- Only output measurement queries, and no validation queries.

- Format each query as a numbered list with no new lines between the queries.

- If there are no relevant queries, output an empty string "".
\end{lstlisting}
\end{tcolorbox}

\newpage 

\begin{tcolorbox}[title=Generate Code from Queries Prompt, mypromptbox, verbatim]
\footnotesize
\begin{lstlisting}
You are an intelligent code generation agent responsible for transforming user queries into 
executable Python code using a specified API reference. 

**Task**: When given a query, generate Python code that addresses the query using the provided API.

**Output Requirements**:

- The generated code must be encapsulated in a function named 'get_result,' which takes a 
parameter 'cxr_image' of type 'CXRImage.'

- If the method 'find' returns an empty list, the code must return the object that is not 
found like this: "[object_name] not present in the image."

- After executing the code, return a string that answers the original query based on the 
results obtained.

**Code Template**: Use the following structure for your code:

```python
def get_result(cxr_image: CXRImage):
    # Implement the logic to process the CXRImage and solve the query
    return result  # Replace with the actual result based on the execution
```

**API Reference**: 

{api_reference}
\end{lstlisting}
\end{tcolorbox}

\begin{tcolorbox}[title=API Reference, mypromptbox, verbatim]
\scriptsize
\begin{lstlisting}
class CXRObject:
    """A Python class containing an image object with bounding box information.
    Parameters
    ----------
    object_name : str
        The object name.
    bbox : Tuple[float, float, float, float]
        A tuple representing the bounding box in the format (left, lower, right, upper).
    The object is considered a point if left == right and lower == upper.
    Methods
    -------
    is_point() -> bool
        Returns true if the objects bounding box is a point.
    get_center() -> Tuple[float, float]
        Returns the center of the object's bounding box or the point if it's a single point.
    """

class CXRImage:
    """A Python class containing an image related to a report as well as relevant information.
    Parameters
    ----------
    rid: str
        Id of report object
    reports: dict
        Reports of "GroundTruth" and models.
    image_path: str
        Full image path.
    original_size: List[int]
        The original WxH of the image.
    pixel_spacing: Tuple[float, float]
        Pixel spacing (mm) in the x- and y-directions.
    cache: ModuleCache
        An instance of ModuleCache to retrieve precomputed segmentation and measurement data.
    Methods
    -------
    exists(object_name: str) -> bool
        Returns True if the object is found in the image, and False otherwise.
    find(object_name: str) -> List[CXRObject]
        Returns a list of CXRObjects found in the image matching the object_name.
    segment(object_name: str) -> CXRSegmentation
        Returns a segmentation map of the image based on the object specified.
    within(ob: CXRObject, region: CXRSegmentation) -> bool
        Returns true of if the object center is within the region.
    distance(obj_a: CXRObject  obj_b: CXRObject) -> float
        Returns the distance (in cm) between the center of two objects in the image.
    diameter(obj: CXRObject) -> float
        Returns the diameter (in cm) of the CXRObject.
    dimensions(obj: CXRObject) -> Tuple[float, float]
        Returns the dimensions (in cm) of the CXRObject according to major axis x minor axis.
    width(segmentation: CXRSegmentation) -> float
        Returns the greatest width (in cm) of a segmentation.
    height(segmentation: CXRSegmentation) -> float
        Returns the greatest height (in cm) of a segmentation.
    filter(objects: [CXROBject], region: CXRSegmentation) -> List[CXRObject]
        Returns all objects with their center within the region.
    """

class CXRSegmentation:
    """A Python class containing an anatomical region.
    Parameters
    ----------
    object_name : str
        The object name.
    segmentation_map : List[int, int]
        A binary segmentation map of the anatomical region.
    Methods
    -------
    get_pixel_width() -> float
        Returns the widest pixel width of the segmentation
    get_pixel_height() -> float
        Returns the tallest pixel height of the segmentation
    """
\end{lstlisting}
\end{tcolorbox}

\begin{tcolorbox}[title=Update Report Prompt, mypromptbox, verbatim]
\footnotesize
\begin{lstlisting}
You are a highly skilled radiologist responsible for updating an existing radiology report 
based on a list of new results.

**Task**: Using the provided list of new results, carefully revise the original report to 
reflect the new information.

**Requirements**:

- You must use all results in the list.

- If any of the new results contradict details in the original report, update those details 
to align with the new results.

- Retain the formatting, tone, and language of the original report as much as possible.

- If the new results indicate that any sentence or phrase in the original report is 
incorrect, omit it from the updated version.

- If the results contain no new measurements, you must keep the measurements in the 
original report.

**Output Format**:

Return the updated report using the following format: 

Updated report: "{updated-report-here}"
\end{lstlisting}
\end{tcolorbox}

\begin{tcolorbox}[title=Extract ETT Mentions Prompt, mypromptbox, verbatim]
\footnotesize
\begin{lstlisting}
You are an experienced radiologist responsible for finding information on endotracheal 
tube (ET) placement in radiology reports. Your performance and accuracy are crucial for
our patient care quality.

To solve this task, perform the following steps:

1. Identify ET Tube Present: Determine if the report explicitly states that an ET tube 
is present. Note that mentions of ET tube removal or patient extubation indicate that 
the ET tube is no longer present.

2. Extract ET Tube Measurement: If an ET tube is present, extract its relative distance 
to the carina in centimeters (cm) if specified. Positive values indicate placement above 
the carina, while negative values indicate placement below the carina.

3. Determine ET Tube Placement: If an ET tube is present, determine if the report deems 
the placement correct or incorrect. If incorrect, categorize the placement as "too low" 
or "too high," if possible.

If you cannot extract a specific category, use "null". There is no need to guess or invent 
a value.

Adhere to the following rules:

- Interpret measurements such as "less than [x] cm" as "[x] cm".

- Interpret measurements such as "[x]-[y] cm" as the higher value, outputting "[y] cm"

- Interpret measurements such as "2. 0 cm" as "2.0 cm"

- Interpret terms like "stable" and "unchanged" as "correct."

- If the report does not clarify whether the measurement is above or below the carina, 
assume it is above and provide a positive value.

- If you do not find a specific measurement in centimeters (cm) or millimeters (mm), 
never infer or approximate a value, simply output 'null'. This is true even if the report 
specifies anatomical landmarks.

First, write one sentence describing how you solved the task for each step. Finally, format 
your results as a JSON object using the following schema:
```json
{
    'ET_present': bool,
    'ET_measurement': float or null,
    'ET_placement': str or null,
}
```
\end{lstlisting}
\end{tcolorbox}

\begin{tcolorbox}[title=Extract Measurement Mentions Prompt, mypromptbox, verbatim]
\footnotesize
\begin{lstlisting}
You are an experienced radiologist responsible for finding information on object measurements 
in radiology reports. Your performance and accuracy are crucial for our quality of patient care.

To solve this task, perform the following steps:

1. Identify measured object(s): list all objects that the reports include measurements for 
using concrete measurements in centimeters (cm) and/or millimeters (mm). 

Adhere to the following rules:

- Do not include objects specified in qualitative descriptions or anatomical landmarks, such 
as "incorrect position" and "projects into the stomach."

First, write one sentence describing how you solved the task for each step. Finally, format 
your answer as a comma-separated string, following this format:

Reasoning: [1 sentence explaining your reasoning]
Answer: object 1, object 2, etc., or an empty string if no objects are measured.

**Examples:**

Report: The tracheostomy tube ends 3.5 cm from the carina. There is a small apical right 
pneumothorax. Heart size is normal-the endotracheal tube projects into the T2 region.
Reasoning: Only the tracheostomy tube contains a specific measurement in centimeters or millimeters.
Answer: tracheostomy tube

Report: Ill-defined nodule in the right upper lung measuring 1.3 x 1.4 cm. The endotracheal tube 
tip now measures approximately 4.6 cm above the carina-the tip of the right internal 
jugular vein catheter projects over the cavoatrial junction.
Reasoning: Both the nodule and the endotracheal tube tip include specific measurements in 
centimeters or millimeters, which is why the answer consists of both objects but not the 
right internal jugular vein.
Answer: nodule, endotracheal tube
\end{lstlisting}
\end{tcolorbox}


\clearpage

\subsection{Baseline Models}

\noindent \textbf{CheXagent~\cite{chen2024chexagent}.} CheXagent is an instruction-tuned foundation model specifically designed for chest X-ray interpretation. The model consists of a vision encoder for representing CXR images, and a network to bridge the vision and language modalities. This model is trained on CheXinstruct, a large-scale instruction-tuning dataset curated from 28 publicly-available datasets.

\noindent \textbf{CheXpertPlus~\cite{chambon2024chexpert}.} CheXpertPlus, introduced in the CheXpert Plus paper, utilizes a Swinv2~\cite{liu2022swin} architecture with a two-layer BERT decoder~\cite{kenton2019bert} for medical report generation.

\noindent \textbf{GPT-4V~\cite{yang2023dawn}.} GPT-4V (GPT-4 with vision) is a multimodal LLM released by OpenAI, which enables users to instruct GPT-4 to analyze image inputs provided by the user. In our evaluation, we used the API of model ``gpt4o05132024'' and followed the official evaluation protocols to assess its performance. The prompt we used is ``You are a helpful assistant. Please generate a report for the given images, including both findings and impressions. Return the report in the following format: Findings: \{\} Impression: \{\}. ''. 

\noindent \textbf{LLM-CXR~\cite{lee2023llm}.} LLM-CXR is a multimodal large language model that utilizes VQ-GAN to tokenize images, integrating both image and text tokens as input to its base LLM architecture. This model enables CXR-to-report generation, report-to-CXR generation, and CXR-related VQA.

\noindent \textbf{RGRG~\cite{tanida2023interactive}.} RGRG (Region-Guided Radiology Report Generation) employs object detection to extract localized visual features from 29 anatomical regions in chest X-rays. It uses binary classifiers to select salient features and encode abnormalities, followed by a language model generating sentences for each selected region. RGRG was trained on the Chest ImaGenome dataset~\cite{wu2021chest}. 

\noindent
\textbf{MAIRA-2~\cite{bannur2024maira}.} MAIRA-2 is a large multimodal model that combines a radiology-specific image encoder with a Large Language Model (LLM), trained for grounded report generation from chest X-rays. For input, the model accepts X-ray images along with indication, comparison, and technique information. 
For studies containing both frontal and lateral views, we input the technique that ``PA and lateral views of the chest were obtained.''. For studies with only frontal views, we use ``PA view of the chest was obtained.''.

\noindent \textbf{MedVersa~\cite{zhou2024generalist}.} MedVersa is a compound medical AI system that can coordinate multimodal inputs, orchestrate models and tools for varying tasks, and generate multimodal outputs. MedVersa was trained on the MIMIC-CXR train and valid dataset for medical report generation tasks.

\noindent \textbf{RadFM~\cite{wu2023towards}.} RadFM is a versatile radiology foundation model trained on large-scale multi-modal datasets. It supports both 2D and 3D scans, multi-image input, and visual-language interleaving cases. The model's training included the MIMIC-CXR dataset.

\noindent \textbf{RaDialog~\cite{pellegrini2023radialog}.}  RaDialog is a large vision-language model for radiology report generation and interactive dialogue. It integrates visual image features and structured pathology findings with a large language model (LLM), adapted to radiology using parameter-efficient fine-tuning. RaDialog was trained on the MIMIC-CXR.

\noindent
\textbf{RGRG~\cite{tanida2023interactive}.} RGRG (Region-Guided Radiology Report Generation) employs object detection to extract localized visual features from 29 anatomical regions in chest X-rays. It uses binary classifiers to select salient features and encode abnormalities, followed by a language model generating sentences for each selected region. RGRG was trained on the Chest ImaGenome dataset~\cite{wu2021chest}.

\noindent \textbf{VLCI~\cite{chen2023cross}.} VLCI~(Visual-Linguistic Causal Intervention) combines Visual linguistic pre-training using a multiway transformer for cross-modal alignment with Visual-linguistic causal intervention, integrating a pre-trained transformer and Visual and linguistic de-confounding Modules to mitigate cross-modal bias through local and global visual sampling and linguistic estimation using a vocabulary dictionary and visual features.

\subsection{CarinaNet Fine-tuning}
We fine-tuned CarinaNet to develop CarinaNet+, leveraging a private dataset of 1,100 chest X-rays collected from intensive care units across 22 hospitals. The dataset was split into 770 images for training and 330 for validation, with final testing performed on the MIMIC-CXR test set (Table \ref{tab:carinanet_performance}). Training was conducted using the AdamW optimizer \cite{loshchilov2017decoupled} with OneCycleLR scheduling~\cite{smithSuperConvergenceVeryFast2018}. The hyperparameters were configured with a batch size of 32, initial learning rate of 7.28e-5, weight decay of 0.044323, maximum learning rate of 3.74e-4, and percentage start of 0.48062. The model was trained for 1,000 total steps with early stopping patience of 6 epochs.

\clearpage
\subsection{Additional Pipeline Metrics}

    \begin{table*}[htbp]
        \centering
        \small
        \caption{Performance statistics for the tasks of ETT Presence, Measurement, and Placement using the \textbf{original} reports.}
        \begin{tabular}{lcccccccccc}
        \toprule
        & \multicolumn{4}{c}{ETT Presence} & \multicolumn{2}{c}{ETT Measurement} & \multicolumn{4}{c}{ETT Placement} \\
        \cmidrule(lr){2-5}\cmidrule(lr){6-7}\cmidrule(lr){8-11}
        Model & Precision & Recall & F1 & BACC & MAE & MSE & Precision & Recall & F1 & BACC \\
        \midrule
    CheXagent & 0.70 & 0.36 & 0.48 & 0.67 & 1.98 & 9.44 & 0.85 & 1.00 & 0.92 & 0.50 \\
CheXpertPlus & 0.64 & 0.76 & 0.69 & 0.86 & 1.44 & 3.47 & 0.90 & 0.67 & 0.77 & 0.70 \\
Cvt2distilgpt2 & 0.71 & 0.53 & 0.60 & 0.75 & 3.82 & 101.58 & 0.73 & 0.91 & 0.81 & 0.48 \\
GPT4V & 0.20 & 0.22 & 0.21 & 0.57 & 2.35 & 7.35 & 1.00 & 0.50 & 0.67 & 0.50 \\
LLM-CXR & 0.08 & 0.53 & 0.13 & 0.49 & 2.41 & 10.38 & 0.74 & 0.50 & 0.60 & 0.46 \\
MAIRA-2 & 0.63 & 0.82 & 0.71 & 0.89 & 1.43 & 3.91 & 0.76 & 0.97 & 0.85 & 0.50 \\
MedVersa & 0.73 & 0.72 & 0.72 & 0.85 & 1.07 & 2.29 & 0.84 & 0.94 & 0.89 & 0.60 \\
RGRG & 0.50 & 0.84 & 0.63 & 0.88 & 1.58 & 5.44 & 0.77 & 0.94 & 0.85 & 0.50 \\
RaDialog & 0.67 & 0.70 & 0.69 & 0.84 & 0.99 & 1.54 & 0.81 & 0.91 & 0.85 & 0.60 \\
RadFM & 0.28 & 0.05 & 0.08 & 0.52 & 0.65 & 0.64 & 1.00 & 1.00 & 1.00 & 1.00 \\
VLCI & 0.25 & 0.12 & 0.16 & 0.54 & 3.50 & 58.02 & 0.80 & 0.89 & 0.84 & 0.44 \\

        \bottomrule
        \end{tabular}
        \label{tab:ett_metrics_original}
    \end{table*}

    \begin{table*}[htbp]
        \centering
        \small
        \caption{Performance statistics for the tasks of ETT Presence, Measurement, and Placement using the \textbf{updated} reports.}
        \begin{tabular}{lcccccccccc}
        \toprule
        & \multicolumn{4}{c}{ETT Presence} & \multicolumn{2}{c}{ETT Measurement} & \multicolumn{4}{c}{ETT Placement} \\
        \cmidrule(lr){2-5}\cmidrule(lr){6-7}\cmidrule(lr){8-11}
        Model & Precision & Recall & F1 & BACC & MAE & MSE & Precision & Recall & F1 & BACC \\
        \midrule
    CheXagent & 0.76 & 0.36 & 0.49 & 0.68 & 0.68 & 0.93 & 0.90 & 0.93 & 0.91 & 0.66 \\
CheXpertPlus & 0.68 & 0.76 & 0.72 & 0.86 & 0.66 & 0.76 & 0.94 & 0.88 & 0.91 & 0.84 \\
Cvt2distilgpt2 & 0.72 & 0.53 & 0.61 & 0.75 & 0.89 & 2.92 & 0.88 & 0.82 & 0.85 & 0.76 \\
GPT4V & 0.58 & 0.22 & 0.32 & 0.60 & 0.39 & 0.16 & 1.00 & 1.00 & 1.00 & 1.00 \\
LLM-CXR & 0.58 & 0.53 & 0.55 & 0.75 & 0.91 & 1.95 & 0.97 & 0.82 & 0.89 & 0.87 \\
MAIRA-2 & 0.68 & 0.82 & 0.74 & 0.89 & 0.99 & 3.05 & 0.91 & 0.83 & 0.87 & 0.80 \\
MedVersa & 0.80 & 0.72 & 0.76 & 0.85 & 0.86 & 1.98 & 0.94 & 0.85 & 0.89 & 0.81 \\
RGRG & 0.60 & 0.84 & 0.70 & 0.89 & 0.76 & 1.10 & 0.96 & 0.90 & 0.93 & 0.88 \\
RaDialog & 0.69 & 0.70 & 0.70 & 0.84 & 0.86 & 2.36 & 0.92 & 0.84 & 0.88 & 0.81 \\
RadFM & 0.47 & 0.05 & 0.09 & 0.52 & 1.04 & 3.49 & 1.00 & 0.80 & 0.89 & 0.80 \\
VLCI & 0.54 & 0.12 & 0.20 & 0.56 & 0.98 & 1.96 & 0.88 & 0.78 & 0.82 & 0.64 \\

        \bottomrule
        \end{tabular}
        \label{tab:ett_metrics_updated}
    \end{table*}

\clearpage
\subsection{Robustness Analysis}

We stratify our analysis by patient demographics (sex, age groups) and clinical contexts (presence of comparison studies, clinical indications). We filter for the most common clinical indications when an endotracheal tube is present and categorize them into respiratory, intubation, or other. We report the average performance of 11 models in Table \ref{tab:ett_metrics_stratified}.

    \begin{table}[htbp]
    \centering
    \small
    \caption{Performance for the tasks of ETT Presence, Measurement, and Placement using the updated reports stratified by clinical context.}
    \vspace{-8pt}
    \setlength{\tabcolsep}{2pt}  
    \begin{tabular}{lcccccccccccc}
    \toprule
    & \multicolumn{3}{c}{Gender} & \multicolumn{3}{c}{Age} & \multicolumn{2}{c}{Comparison} & \multicolumn{3}{c}{Indication} & All \\
    & Female & Male & Unknown & 0-59 & 60+ & Unknown & True & False & Respiration & Intubation & Other & All \\
    \midrule
    Presence Precision & 0.62 & 0.62 & 0.78 & 0.66 & 0.57 & 0.78 & 0.72 & 0.60 & 0.67 & 0.74 & 0.59 & 0.64 \\
Measurement MAE & 0.64 & 0.72 & 1.15 & 0.60 & 0.90 & 1.15 & 0.73 & 0.78 & 0.95 & 0.65 & 0.87 & 0.82 \\
Placement Precision & 0.96 & 0.82 & 0.73 & 0.95 & 0.89 & 0.73 & 0.76 & 0.94 & 0.84 & 0.83 & 0.95 & 0.94 \\

    \bottomrule
    \end{tabular}
    \label{tab:ett_metrics_stratified}
    \end{table}
    
\noindent We validate our approach on the CheXpert Plus (Stanford-CXR) test set. The average results of 11 models are summarized in Table \ref{tab:ett_metrics_chexpert}. However, note that the number of cases with endotracheal tubes is significantly fewer than for MIMIC-CXR 2.0.0, with overlapping model predictions and ground truths only ranging from 4-18 reports.

\begin{table}[!htb]
\centering
\small
\caption{Average performance for the tasks of ETT Presence, Measurement, and Placement using the updated reports of CheXpert Plus}
\vspace{-8pt}
\setlength{\tabcolsep}{4pt}
\begin{tabular}{lcccccc}
\toprule
\multicolumn{2}{c}{ETT Presence (Precision)} & \multicolumn{3}{c}{ETT Measurement (Composite)} & \multicolumn{2}{c}{ETT Placement (Precision)} \\ Original & Updated & Original & Updated & Improvement & Original & Updated \\ \midrule 
 0.52 & \textbf{0.68} & 5.05 & \textbf{2.03} & 149.0\% & 0.58 & \textbf{0.68} \\
\bottomrule
\end{tabular}
\label{tab:ett_metrics_chexpert}
\end{table}
    
\noindent We conduct an ablation study using MedKLIP backbones for the \textsc{ResNet-50+} endotracheal tube detection module. Since the original ImageNet backbone outperforms MedKLIP (Table \ref{tab:resnetmedklip}), we use ImageNet in the final pipeline.

\begin{table}[!htb]
\centering
\small
\caption{Performance of the \textsc{ResNet-50+} Module using ImageNet and MedKLIP backbones.}
\vspace{-8pt}
\setlength{\tabcolsep}{7pt}
\begin{tabular}{lcccccc}
\toprule
 Pretraining &  ACC & BACC & F1 & Prec. & Rec. & AUC \\ 
 \midrule
ImageNet & \textbf{0.94} & \textbf{0.97} & \textbf{0.73} & \textbf{0.57} &\textbf{ 0.99} & \textbf{0.97} \\
MedKLIP & 0.91 & 0.94 & 0.63 & 0.46 & \textbf{0.99} & 0.94 \\
\bottomrule
\end{tabular}
\label{tab:resnetmedklip}
\end{table}

\noindent Table \ref{tab:ett_run_all_mimic} shows the performance of running the FactCheXcker pipeline on all reports regardless of the mention of ETT. Although this approach improves false negatives, the precision score significantly drops.



\begin{table*}[htbp]
\centering
\small
\caption{Performance comparison between the Original Model and the Updated Model with the FactCheXcker modules applied to all reports, regardless of ETT mention.}
\setlength{\tabcolsep}{5pt}
\vspace{-8pt}
\begin{tabular}{lcccccccccc}
\toprule
& \multicolumn{2}{c}{ETT Presense} & \multicolumn{6}{c}{ETT Measurement} & \multicolumn{2}{c}{ETT Placement} \\
\cmidrule(lr){2-3} \cmidrule(lr){4-9} \cmidrule(lr){10-11}
Model & \multicolumn{2}{c}{Precision} & \multicolumn{3}{c}{MAE} & \multicolumn{3}{c}{Composite} & \multicolumn{2}{c}{Precision} \\
\cmidrule(lr){2-3} \cmidrule(lr){4-6} \cmidrule(lr){7-9} \cmidrule(lr){10-11}
 & Original & Updated & Original & Updated & Improvement & Original & Updated & Improvement & Original & Updated \\
\midrule
CheXagent & \textbf{0.70} & 0.57 & 1.98 & \textbf{0.68} & 191.0\% & 4.12 & \textbf{0.93} & 343.0\% & 0.85 & \textbf{0.90} \\
CheXpertPlus & \textbf{0.64} & 0.57 & 1.44 & \textbf{0.66} & 118.0\% & 2.09 & \textbf{0.90} & 132.0\% & 0.90 & \textbf{0.94} \\
Cvt2distilgpt2 & \textbf{0.71} & 0.57 & 3.82 & \textbf{0.89} & 329.0\% & 6.37 & \textbf{1.22} & 422.0\% & 0.73 & \textbf{0.88} \\
GPT4V & 0.20 & \textbf{0.57} & 2.35 & \textbf{0.39} & 503.0\% & 11.19 & \textbf{0.53} & 2011.0\% & \textbf{1.00} & \textbf{1.00} \\
LLM-CXR & 0.08 & \textbf{0.57} & 2.41 & \textbf{0.91} & 165.0\% & 18.54 & \textbf{1.25} & 1383.0\% & 0.74 & \textbf{0.97} \\
MAIRA-2 & \textbf{0.63} & 0.57 & 1.43 & \textbf{0.99} & 44.0\% & 2.01 & \textbf{1.36} & 48.0\% & 0.76 & \textbf{0.91} \\
MedVersa & \textbf{0.73} & 0.57 & 1.07 & \textbf{0.86} & 24.0\% & 1.49 & \textbf{1.18} & 26.0\% & 0.84 & \textbf{0.94} \\
RGRG & 0.50 & \textbf{0.57} & 1.58 & \textbf{0.76} & 108.0\% & 2.51 & \textbf{1.04} & 141.0\% & 0.77 & \textbf{0.96} \\
RaDialog & \textbf{0.67} & 0.57 & 0.99 & \textbf{0.86} & 15.0\% & 1.43 & \textbf{1.18} & 21.0\% & 0.81 & \textbf{0.92} \\
RadFM & 0.28 & \textbf{0.57} & \textbf{0.65} & 1.04 & -38.0\% & 8.12 & \textbf{1.42} & 472.0\% & \textbf{1.00} & \textbf{1.00} \\
VLCI & 0.25 & \textbf{0.57} & 3.50 & \textbf{0.98} & 257.0\% & 21.88 & \textbf{1.34} & 1533.0\% & 0.80 & \textbf{0.88} \\
\midrule
Average & 0.49 & \textbf{0.57} & 1.93 & \textbf{0.82} & 135.0\% & 7.25 & \textbf{1.12} & 546.0\% & 0.84 & \textbf{0.94} \\
\bottomrule
\end{tabular}
\label{tab:ett_run_all_mimic}
\end{table*}